\documentclass[conference]{IEEEtran}

\IEEEoverridecommandlockouts 

\usepackage{cite}
\usepackage{amsmath,amssymb,amsfonts}
\usepackage{algorithmic}
\usepackage{graphicx}
\usepackage{textcomp}
\usepackage{xcolor}
\def\BibTeX{{\rm B\kern-.05em{\sc i\kern-.025em b}\kern-.08em
    T\kern-.1667em\lower.7ex\hbox{E}\kern-.125emX}}
\usepackage{booktabs}

\usepackage{enumitem}
\usepackage{multirow}
\usepackage[super]{nth}
\usepackage{tikz}

\usetikzlibrary{tikzmark}
\tikzset{circledColor/.style={circle,draw,inner sep=0.1em,line width=0.04em}}
\usepackage{soul}

\usepackage{setspace}
\usepackage[spaces,hyphens]{xurl} 
\usepackage[colorlinks,allcolors=blue]{hyperref}
\usepackage{cleveref}
\usepackage{adjustbox}
\usepackage{makecell}

\usepackage{titlesec}
\titlespacing\section{0pt}{0.6\baselineskip}{0.3\baselineskip}
\titlespacing\subsection{0pt}{0.3\baselineskip}{0.15\baselineskip}
\titlespacing\subsubsection{0pt}{0.2\baselineskip}{0.1\baselineskip}

\title{\LARGE \bf
TinyCL: An Efficient Hardware Architecture for Continual Learning on Autonomous Systems
}

\author{\IEEEauthorblockN{Eugenio Ressa\IEEEauthorrefmark{1}, Alberto Marchisio\IEEEauthorrefmark{2}, Maurizio Martina\IEEEauthorrefmark{1}, Guido Masera\IEEEauthorrefmark{1}, Muhammad Shafique\IEEEauthorrefmark{2}}
\IEEEauthorblockA{\IEEEauthorrefmark{1}Department of Electronics and Telecommunications, Politecnico di Torino, Turin, Italy}
\IEEEauthorblockA{\IEEEauthorrefmark{2}eBrain Lab, Division of Engineering, New York University Abu Dhabi, Abu Dhabi, UAE} 
\IEEEauthorblockA{\small{Emails: s281642@studenti.polito.it, \{alberto.marchisio, muhammad.shafique\}@nyu.edu, \{maurizio.martina, guido.masera\}@polito.it}}\\
\vspace*{-30pt}}

\usepackage{fancyhdr,lipsum}
\fancypagestyle{firstpage}{
  \fancyhf{}
  \fancyhead[C]{To appear at the 2025 International Joint Conference on Neural Networks (IJCNN), Rome, Italy, July 2025.}
  \fancyfoot[C]{\thepage}
}

\begin{document}

\maketitle
\thispagestyle{firstpage}

\begin{abstract}

The Continuous Learning (CL) paradigm consists of continuously evolving the parameters of the Deep Neural Network (DNN) model to progressively learn to perform new tasks without reducing the performance on previous tasks, i.e., avoiding the so-called catastrophic forgetting. However, the DNN parameter update in CL-based autonomous systems is extremely resource-hungry. The existing DNN accelerators cannot be directly employed in CL because they only support the execution of the forward propagation. Only a few prior architectures execute the backpropagation and weight update, but they lack the control and management for CL. Towards this, we design a hardware architecture, TinyCL, to perform CL on resource-constrained autonomous systems. It consists of a processing unit that executes both forward and backward propagation, and a control unit that manages memory-based CL workload. To minimize the memory accesses, the sliding window of the convolutional layer moves in a snake-like fashion. Moreover, the Multiply-and-Accumulate units can be reconfigured at runtime to execute different operations. As per our knowledge, our proposed TinyCL represents the first hardware accelerator that executes CL on autonomous systems. We synthesize the complete TinyCL architecture in a 65 nm CMOS technology node with the conventional ASIC design flow. It executes 1 epoch of training on a Conv + ReLU + Dense model on the CIFAR10 dataset in 1.76 s, while 1 training epoch of the same model using an Nvidia Tesla P100 GPU takes 103 s, thus achieving a $58 \times$ speedup, consuming 86 mW in a 4.74 mm$^2$ die.

\end{abstract}

\section{Introduction}

In recent years, Deep Neural Networks (DNNs) have been deployed in several applications, like computer vision, finance, healthcare, and robotics~\cite{Chai_2021MLWA_DLComputerVision, Chandani_2022IJDATS_MLFinance, An_2023Sensors_MLHelathcare, Soori_2023COGR_DLRobotics}. A common practice is to train a DNN on the desired task using the training set, and then deploy the trained model on the target for inference~\cite{Capra_2020Access_SurveyDNNs}. In this way, it is possible to conduct DNN training on large data centers (e.g., using high-end GPUs), and then conduct resource-constrained optimizations (e.g., compression) to deploy DNN inference on autonomous systems~\cite{Dhilleswararao_2022Access_EfficientDL, marchisio2021feeca, marchisio2023swifttron, putra2024snn4agents}. However, this practice limits the dynamic capabilities of DNNs that cannot adapt to new tasks or a distribution variation of the input data within the same task. In this regard, the Continual Learning (CL) paradigm enables the dynamic change of DNN parameters to evolve and learn new tasks (or new classes)~\cite{Wang_2023arxiv_CLSurvey, aswani2024examining, minhas2024continual}. The main goal of CL-based algorithms is to avoid Catastrophic Forgetting (CF), i.e., the DNN should maintain the knowledge of how to perform the previous tasks while learning new tasks~\cite{Mccloskey_1989Elsevier_CatastrophicForgetting}.

\subsection{Target Research Problem}

Since they require the execution of the backpropagation and parameters' update, CL algorithms impose more demanding computation and memory resources than traditional systems that conduct only inference~\cite{Chen_2020Engineering_SurveyDNNAccelerators}. Hence, it is extremely important to execute CL algorithms in an efficient manner to be able to deploy them onto resource-constrained autonomous systems.

Most of the existing architectures and optimizations for DNNs focus only on optimizing the inference process~\cite{Capra_2020MDPI_UpdatedSurveyDNN}. Therefore, they cannot execute the backward operations required by CL algorithms. Some architectures~\cite{Lee_2021SSCS_HWAcceleratorTraining} that have been proposed to accelerate the training can be adopted to conduct backpropagation computations, but they do not support the execution of CF-avoiding policies.





\subsection{Novel Contributions}

To overcome these limitations, we propose TinyCL, a hardware architecture that can efficiently execute CL operations on autonomous systems. Our architecture reuses the same processing units for computing the forward and backward computations, and a specialized control unit dictates the data flow based on the CL policy. In a nutshell, our contributions are (see \Cref{fig:Novel_contributions}):

\begin{figure}[!t]
  \centering
  \includegraphics[width=\linewidth]{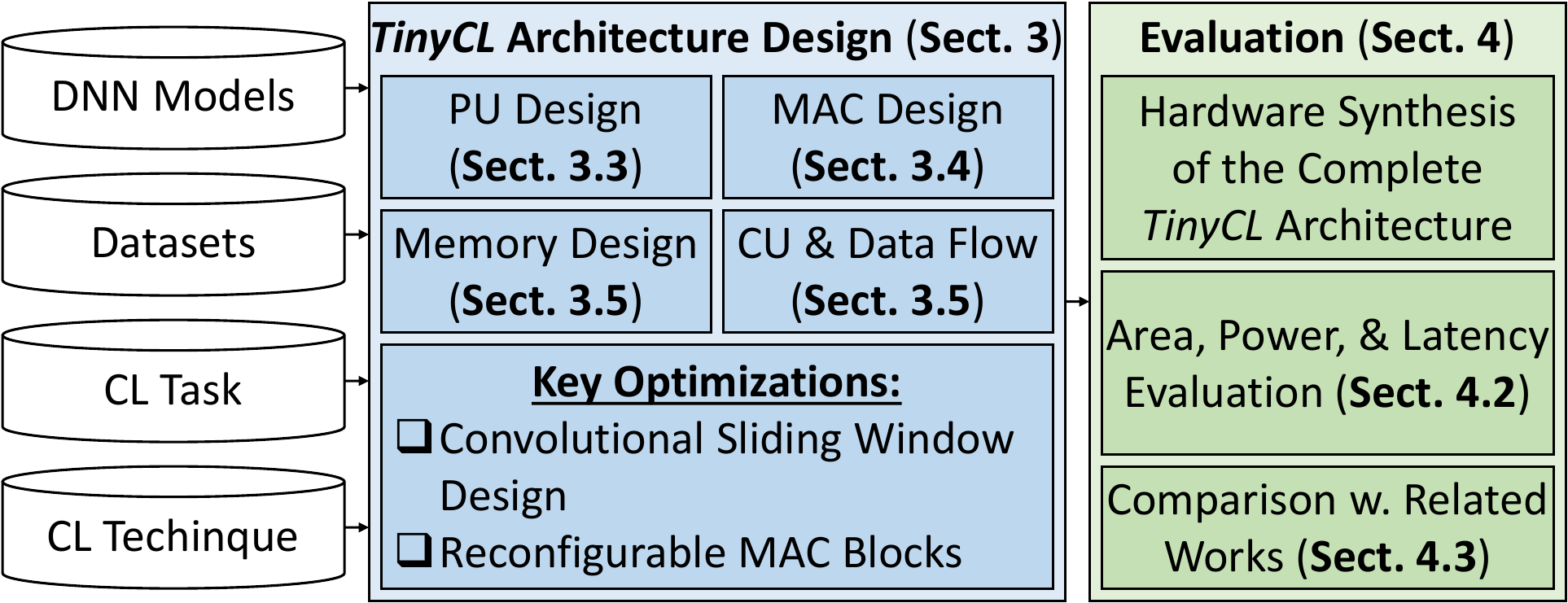}
  \caption{Overview of our novel contributions in this work.}
  \label{fig:Novel_contributions}
\end{figure}

\begin{itemize}[leftmargin=*]
    \item RTL Design of the complete TinyCL architecture, in which multiple Processing Units execute the computation in parallel and the convolutional sliding window is designed following a snake-like pattern. (\textit{\Cref{sec:TinyCL_Design}})
    \item Synthesis of the TinyCL architecture using the conventional ASIC flow for a 65 nm CMOS technology node. (\textit{\Cref{sec:experiments}})
    \item Compared to its equivalent software-level implementation on an Nvidia Tesla P100 GPU, the TinyCL architecture achieves $35 \times$ speedup; compared to other DNN training accelerators in the literature, the TinyCL architecture achieves lower latency, power consumption, and area, thus making it suitable for being adopted on resource-constrained autonomous systems. (\textit{\Cref{subsec:comparison}})
\end{itemize}

\section{Background and Related Work}
\subsection{Convolutional Neural Networks}
Convolutional Neural Networks (CNNs) are a branch of DNNs used in image recognition tasks.
Convolutional and dense layers are the basis blocks of every CNN.
In a convolutional layer, a multi-channel input feature $V$ is convoluted with a 4D kernel $K$ to produce a multi-channel output $Z$ (see \Cref{eq:Z}).

\begin{equation}
\label{eq:Z}
\begin{gathered}
Z(i,j,k)= c(K,V,s)_{i,j,k} = \\
\sum_{l,m,n} [V(l,(j-1) \cdot s+m,(k-1) \cdot s+n) \cdot K(i,l,m,n)]
\end{gathered}
\end{equation}

Applying the Stochastic Gradient Descent (SGD), the computation to propagate the gradient across layers is a convolution between the previous kernel $K$ and the gradient $G$ propagated from the previous layer, as shown in \Cref{eq:h}, while the gradient of the kernel is computed through \Cref{eq:gradient_kernel}.

\begin{equation}
\label{eq:h}
\begin{gathered}
h(K,G,s)_{i,j,k} =\frac{\partial}{\partial V_{i,j,k}} J(V,K) = \\
\resizebox{0.91\hsize}{!}{$\sum_{l,m\ s.t. (l-1) \cdot s+m=j} \left( \sum_{n,p\ s.t. (n-1) \cdot s+p=k} \left( \sum_{q} G_{q,l,n} \cdot K_{q,i,m,p} \right) \right)$}
\end{gathered}
\end{equation}

\begin{equation}
\label{eq:gradient_kernel}
\begin{gathered}
g(G,V,s)_{i,j,k,l} = \frac{\partial}{\partial K_{i,j,k}} J(V,K) = \\
\sum_{m,n} G_{i,m,n} \cdot V_{j,(m-1) \cdot s+k,(n-1) \cdot s+l}
\end{gathered}
\end{equation}

The Dense layer computes a matrix multiplication between a 1D row matrix $I$ with a 2D weight matrix $W$ (see \Cref{eq:y_n}).

\begin{equation}
\label{eq:y_n}
y_n = \sum_{i = 0}^{m} I_i \cdot W_{i,n}   
\end{equation}

Applying SGD, we can compute the gradient propagation that is a matrix multiplication between the input gradient $dY$ propagated from the previous layer and the transpose of the weight (see \Cref{eq:dX}).

\begin{equation}
\label{eq:dX}
dX_{i} = \sum_{n=0}^{N}dY_n \cdot W^{T}_{n,i}
\end{equation}

Then, the gradient of the weights is computed as in \Cref{eq:dW}.

\begin{equation}
\label{eq:dW}
dW_{i,n} = I_{i} \cdot dY_n  
\end{equation}

\subsection{Continual Learning Algorithms}


The goal of CL, also known as incremental learning, lifelong learning, or sequential learning, is to gradually learn from different data streams and extend the acquired knowledge~\cite{Thrun_1995RAS_LifelongRobotLearning}. The data streams can be associated with different tasks that the CL system can perform. Incrementally learning all the tasks is an NP-hard problem~\cite{Knoblauch_2020ICML_ContinualLearningNPhard}. Moreover, the major issue of CL is represented by CF, i.e., the ability of the system to perform previously learned tasks degrades over time when new tasks are added~\cite{Mccloskey_1989Elsevier_CatastrophicForgetting}. In the literature, several CL methods have been proposed to mitigate CF. CL algorithms can be categorized as regularization-based methods, memory-based methods, and dynamic approaches.

Regularization-based methods apply constraints in the weight update phase to mitigate CF. Weight regularization methods such as Elastic Weight Consolidation (EWC)~\cite{Kirkpatrick_2016arxiv_CatastrophicForgettingEWC} impose a quadratic penalty to selectively regularize the parameters based on their importance to perform the previous tasks, calculated through the Fisher information matrix. Function regularization methods like the Learning without Forgetting (LwF)~\cite{Li_2018TPAMI_LwF} employ knowledge distillation to learn the training samples of the new tasks while preserving the knowledge of the previous tasks.

Memory-based approaches, also known as replay-based methods, retrain or finetune the DNN jointly using samples from previous tasks and samples from new tasks. The Gradient Episodic Memory (GEM) method~\cite{LopezPaz_2017NeurIPS_GEM} constrains the parameter update such that the training loss of each individual previous task does not increase. Its variant, called A-GEM~\cite{Chaudhry_2019ICLR_AGEM}, ensures that the average training loss for all previous tasks does not increase. The Incremental Classifier and Representation Learning (iCaRL)~\cite{Rebuffi_2017CVPR_iCaRL} method stores a subset of training samples for each task and jointly minimizes the training loss for new tasks and the distillation loss for the previously learned tasks. Experience Replay (ER)~\cite{Hayes_2019ICRA_ExperienceReplay} combines training with samples of the new tasks and old samples that are stored in a replay memory. The Maximally Interfered Retrieval (MIR) method~\cite{Aljundi_2019NeurIPS_MIR} selects the samples from the old tasks that would have the largest impact on the forgetting property. The Gradient-based Sample Selection (GSS) strategy~\cite{Aljundi_2019NeurIPS_GSS} maximizes the gradient diversity of the stored sample subset. Greedy Sampler and Dumb Learner (GDumb) approach~\cite{Prabhu_2020ECCV_GDumb} greedily stores training samples in the memory buffer to maintain a balanced class distribution. The replay data can also be generated at runtime by using the Deep Generative Replay (DGR) method~\cite{Shin_2017NeurIPS_DGR}, where a generator creates synthetic samples that contain previous tasks knowledge.

Dynamic approaches dynamically increase the DNN architecture to learn features of new tasks. The Continual Neural Dirichlet Process Mixture (CN-DPM) method~\cite{Lee_2020ICLR_CNDPM} retains the knowledge of the previous tasks by building a mixture of experts where a new model is trained for a new task, while the existing models for the previous tasks are not modified. The Progressive Segmented Training (PST) method~\cite{Du_2019ICMLA_PST} focuses on a single network and, when training on a new task, it divides the parameters into two groups according to their importance to perform that task. The group of important parameters is frozen to preserve the current knowledge, while the other group is saved and can be updated when learning future tasks.

\textit{Our architecture supports memory-based approaches due to the simplicity of their hardware implementation, but it can be easily extended to execute other CL algorithms.}

\subsection{DNN Training Accelerators}


Various hardware architectures implemented in ASIC or FPGA for accelerating DNN training have been proposed in recent years. 
The work in~\cite{Venkataramani_2017ISCA_ScaleDeep} is composed of heterogeneous processing tiles to efficiently execute different operations with diverse computational characteristics. 
The DeepTrain architecture~\cite{Kim_2018TCAD_DeepTrain} deploys heterogeneous programmable data flows to achieve data reuse during different training operations. 
The Gist architecture~\cite{Jain_2018ISCA_Gist} utilizes layer-specific encoding schemes to exploit redundancy in DNN training by storing the feature maps computed during the forward pass and reusing them in the backward pass. 
The SIGMA accelerator~\cite{Qin_2020HPCA_SIGMA} supports irregular sparse workload. 
To efficiently handle sparsity, the Procrustes accelerator~\cite{Yang_2020MICRO_Procrustes} employs a dense tensor dimension for performing arithmetic operations that involve sparse tensors. 
The LNPU architecture~\cite{Lee_2019ISSCC_LNPU} implements fine-grained mixed precision to perform training. 
The HNPU architecture~\cite{Han_2021JSSC_HNPU} supports low-precision training by dynamically configuring the fixed-point representation.
The FlexBlock architecture~\cite{Noh_2023TC_FlexBlock} supports multiple block floating-point precisions. 
The ETA architecture~\cite{Lu_2023TNNLS_ETA} performs training based on the proposed piecewise integer format. 
The work in~\cite{Kim_2019ISSCC_MobileDepRLAccelerator} trains the DNN through
a two-step process that consists of sample collection and policy update for continuous control of the behavior.
\textit{The above-mentioned architectures allow the execution of standard training in an efficient manner. However, the execution of CL algorithms requires additional abstraction layers to correctly manage the workload.}

\subsection{Optimizations for Continual Learning at the Edge}

Recent works have proposed specialized optimization to improve the efficiency of CL algorithms on resource-constrained devices. 
Huai et al.~\cite{Huai_2023FGCS_ZerorizedBatchNormalization} designed a method for improving the efficiency during the re-training phase of latency-constrained high-end edge systems. 
The SparCL framework~\cite{Wang_2022NeurIPS_SparCL} enables efficient continual learning through weight and gradient sparsity. 
The SIESTA algorithm~\cite{Harun_2023arxiv_SIESTA} proposed to alternate awake cycles with sleep cycles to consolidate the memory. 
The Miro methodology~\cite{Ma_2023MobiCom_Miro} designs the memory hierarchy to leverage the tradeoff between energy and accuracy.

\textit{While these works demonstrated potential to improve the efficiency in some parts of the process, they do not consider the complete system implemented on a specialized accelerator, as we do in this paper.}

\section{TinyCL Architecture Design}
\label{sec:TinyCL_Design}




\subsection{Data quantization}

Since high-precision data is not necessary, a 16-bit fixed-point (4-bit integer + 12-bit fractional data) format is used. Accordingly, the work in~\cite{gaur2020training} suggests that value clipping is useful to increase accuracy in DNNs where batch normalization is not applied.

\subsection{Top-Level Architecture}
\begin{figure}[!ht]
  \centering
  \includegraphics[width=.99\linewidth]{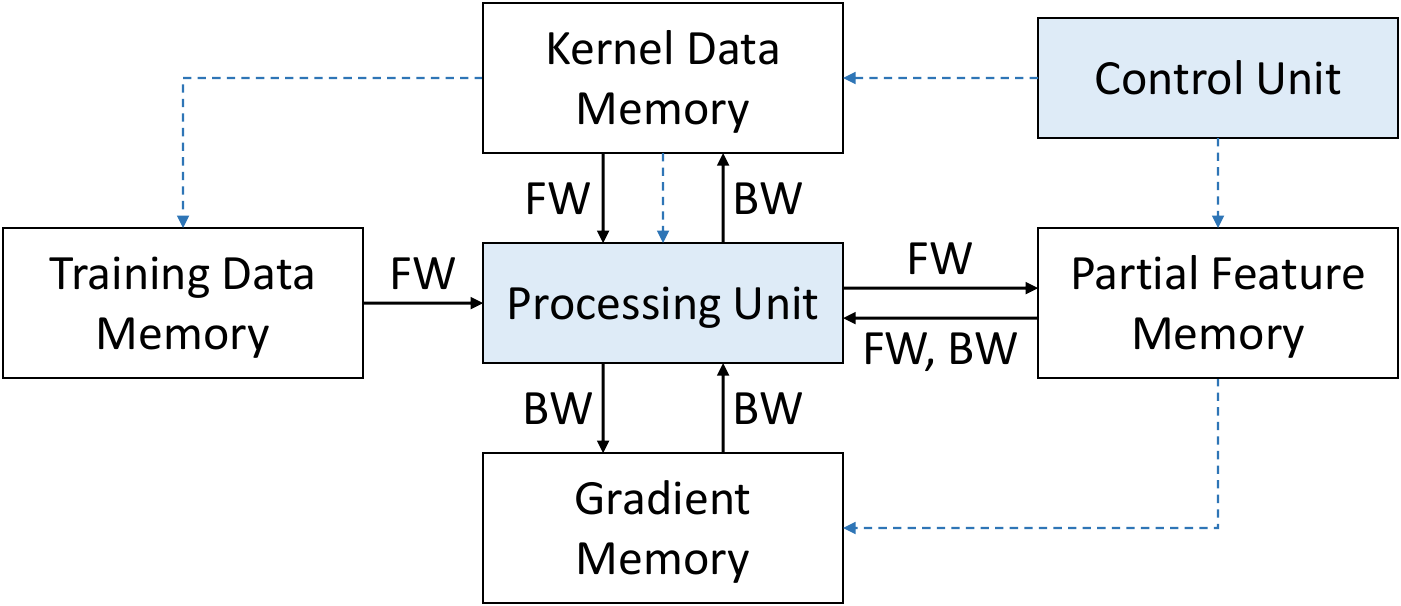}
  \caption{Top-level view of the TinyCL architecture.}
  \label{fig:top_level}
\end{figure}

Our system is composed of 4 groups of data memory ( GDumb memory, Gradient memory, Kernel memory and Partial Feature memory), a control unit and a processing unit, as depicted in \Cref{fig:top_level}.

\subsection{Processing Unit}

\begin{figure}[!ht]
  \centering
  \includegraphics[width=.99\linewidth]{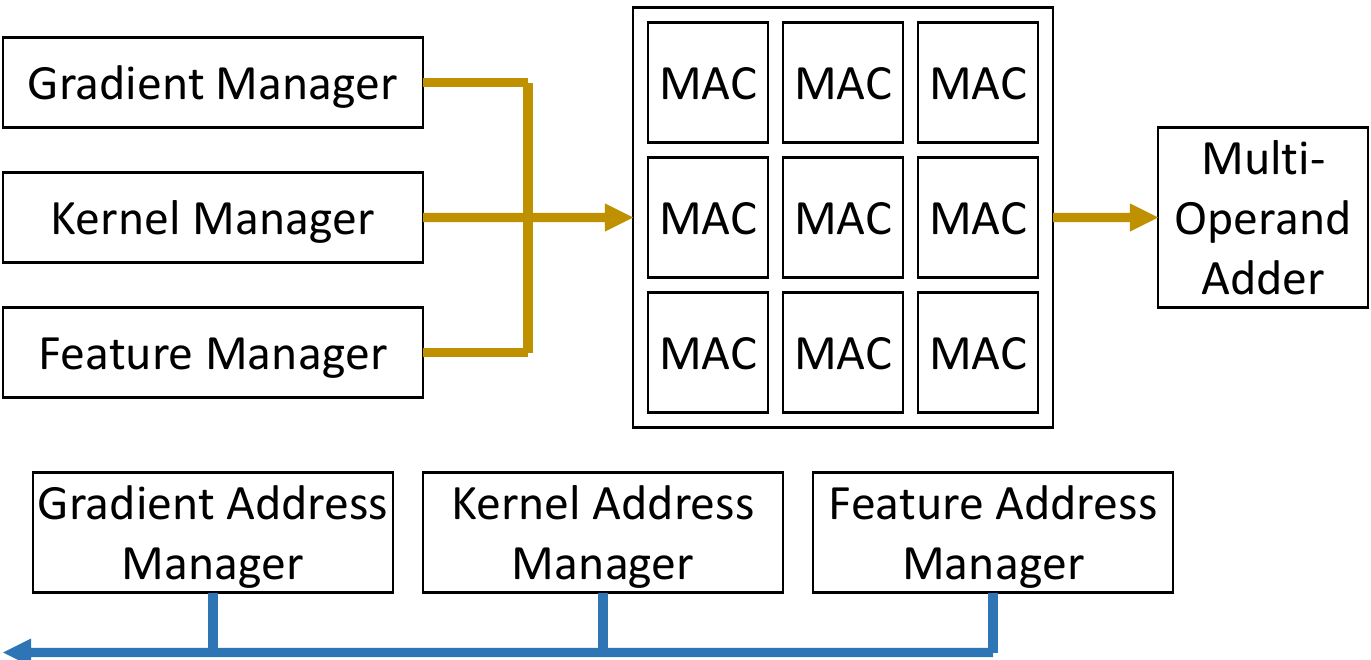}
  \caption{Overview of the processing unit architecture.}
  \label{fig:PU}
\end{figure}

Our processing unit, depicted in \Cref{fig:PU}, includes 9 parallel multiply and accumulate (MAC) blocks, each of them executing an 8-operand multiplication and addition in parallel.
3 manager units (gradient, kernel, and feature) are designed to drive the data flow.
3 address managers (gradient, kernel, and feature) compute the addresses needed to the forward, gradient propagation, and kernel derivative.

\subsection{MAC}

\begin{figure}[!ht]
  \centering
  \vspace{4pt}
  \includegraphics[width=\linewidth]{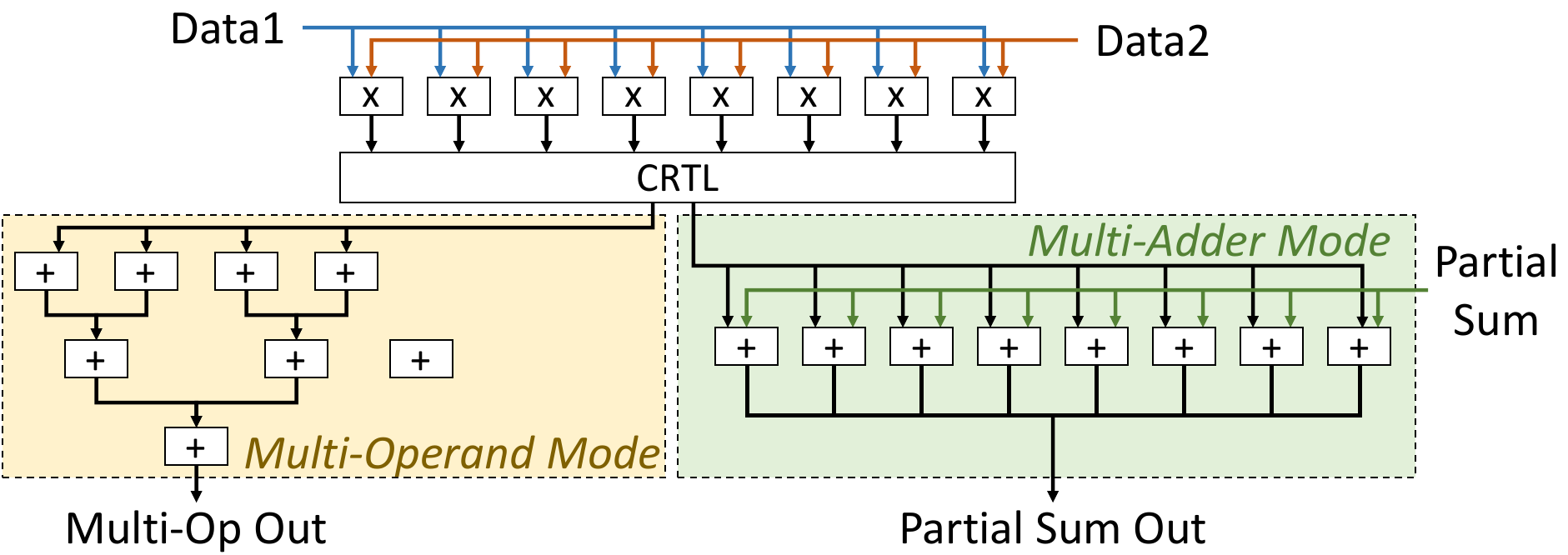}
  \caption{Overview of the multiply-and-accumulate (MAC) architecture.}
  \label{fig:MAC}
\end{figure}

Our MAC is composed of 8 multipliers and 8 adders, as shown in \Cref{fig:MAC}. The multipliers are configured in a parallel fashion, while the adders can be configured in two ways: \textit{multi-operand mode} or \textit{multi-adder mode}.

\begin{itemize}[leftmargin=*]
\item \textbf{Multi-Adder mode}: During kernel gradient calculation, the adders sum the 8 multiplication results with 8 input values to obtain the partial sum outputs. In the weight gradient calculation, as we compute in parallel 8 channels at a time, each time we have to multiply 8 channels of one input feature with one channel of the input gradient. The results of these 8 multiplications will be summed with the previous multiplication done. 
\item \textbf{Multi-Operand mode}: During forward and gradient propagation, we compute 8 input channels at a time. In this case, since the output is the sum of these 8 input channels (3D-convolution), we configure the 7 adders to operate as an adder-tree.
\end{itemize}
To reduce the loss of information, the results of the 16-bit multiplications are kept in full precision and propagated to the 32-bit adders. After the addition, the output is reduced to 16 bit, rounded to nearest.

\subsection{Memory}

A typical DNN accelerator that performs only inference needs to save only its parameters. However, a CL accelerator needs to save other data.

\begin{itemize}[leftmargin=*]

\item \textbf{Training Data Memory}: Memory-based CL methods save old training samples when performing a new training cycle. For this reason, such samples must be saved in a memory that can be updated by replacing some samples of old classes with more samples of new classes. The cardinality of each training sample set must be equal, thus we avoid class imbalance problems.
\item \textbf{Partial Feature Memory}: During inference, each layer applies a function $f(k)$ to the input feature and computes an output feature. During back propagation, the derivative of the output with respect to the weights is a function of the input feature used during forward. For this reason, for each layer with weights, we have to save the input feature during inference, to be used backward.
\item \textbf{Kernel Memory}: Obviously, kernel values must be stored for forward and backward processes.
\item \textbf{Gradient Memories}: To temporarily save the gradients to be used between two computations, a couple of memories must be used. The memories shall be 2 because 1 would not be enough. In a multi-channel convolution operation, the feature we are calculating will overwrite a feature we will need in further calculus.
\end{itemize}

To increase throughput and decrease stalls, we design the memories with a port width of $128$ bits, to read 8 features at a time. To further increase the throughput, the SRAM is organized according to the channel. For example, if we compute convolution with an input feature of $32 \times 32$ with $8$ channels and we output a feature of $32 \times 32$ of $8$ channels, we will have $8$ blocks for partial features of $32 \times 32 \times 16$ bits, $64$ blocks of $3 \times 3 \times 16$ bits to save the kernels and $16$ blocks of $32 \times 32 \times 16$ bits memory for gradient propagation. Moreover, dedicated buffers prefetch data from memory.

\subsection{Control Unit and Data Flow}

Our control unit manages the multi-layer computation, passing the actual matrix input and output sizes to the PU. 
For each layer, the Control Unit (CU) manages the data flow from memories to MACs for executing 6 different computations:
\begin{itemize}
    \item Convolution - forward computation
    \item Convolution kernel - gradient computation
    \item Convolution - gradient propagation
    \item Dense layer - forward
    \item Dense layer - gradient propagation
    \item Dense layer - wight derivative 
\end{itemize}
In each of these computations, data are fetched and loaded into buffers, so that at each clock cycle, the respective features are directly processed by the MACs of the PU. In this way, we are able to maintain a high hardware utilization.

\subsubsection{Forward Computation of the Convolutional Layer}
In the forward operation, each multi-channel input feature must be 3D-convoluted with a 4D kernel to create a single pixel in the new feature, as in \Cref{eq:Z}.


To create more channels in the output feature, more 3D kernels are used. Using $9$ instances of the MAC, we compute in parallel a $8 \times 3 \times 3$ 3D convolution.
To output a single output feature, groups of $8 \times 3 \times 3$ input features (a $3 \times 3$ submatrix for each channel) are convolved with a $8 \times 3 \times 3$ matrix of weights (kernels). If the input feature has more input channels, this operation is repeated.
However, most of the input features that are part of the convolution of $9$ output feature (for each output channel) are used in the next output feature calculations.

\begin{figure}[!ht]
\centering
\vspace{10pt}
\includegraphics[width=.95\linewidth]{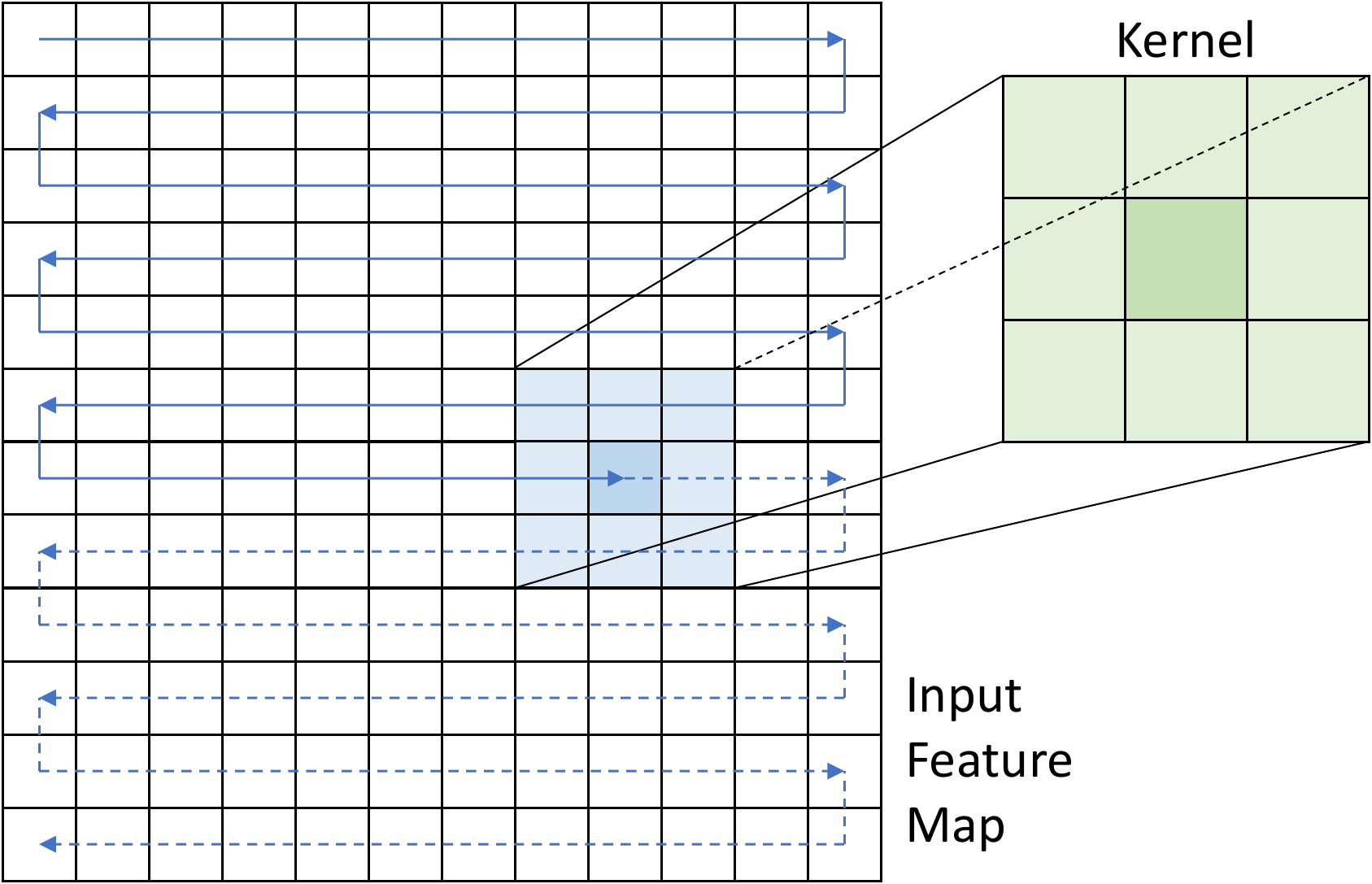}
\caption{Example of forward computation of a channel of a convolutional layer.}
\label{fig:sliding_window_snake}
\end{figure}

When we reach the final feature of a row and move to the next row, we do not restart from column $0$. Instead, we start decreasing the column counter, following a snake-like movement (see \Cref{fig:sliding_window_snake}). In this way, $6$ features are always reused. Hence, when the computations are at full throttle, for each output feature (i.e., for each cycle), we fetch $3$ input features for $8$ channels ($16$-bit each) and write one value. In this way, each cycle, $6$ of the $9$ input features are saved and $3$ new are loaded. 
These $9 \times 8$ values are then sent to the $9$ MACs. Each MAC takes $8$ channels of a feature and multiplies them for $8$ channels of the kernel. These $9$ values are then summed together in a $9$-operand Dadda adder to generate the output. 
This process is conducted for the whole set of output features. 
The addresses to read the input features and the kernels, and to write the new output features are generated by the Forward Address Manager, which uses the dynamic size given by the CU as the bound for the column, row, and channel counters. When the column counter reaches the dynamic size, it will not be zeroed due to snake movement, but its address will be maintained and increased by one row. When the end of the matrix is reached, the channel counter is increased.

%
\subsubsection{Kernel Gradient Computation}
%
%
%
To calculate the gradient of the kernels, a 2D convolution must be executed, as in \Cref{eq:gradient_kernel}. The input gradient is convoluted with $9$ different sub-matrices of the input feature, which are shifted according to the respective position of the gradient kernel. Since the input feature is padded with $2$ stripes of zeros on the right and bottom borders, the resulting matrix has sizes increased by $2$ in both dimensions. 
The index of the kernels is also used for indexing which MAC is used to compute the gradient, according to \Cref{eq:MAC}.

\begin{equation}
\label{eq:MAC}
MAC_{k,l} <= g(G,V,s)_{i,j,k,l} = \frac{\partial}{\partial K_{i,j,k,l}} J(V,K)
\end{equation}

Therefore, the features are shared between convolutions (to an even larger extent than for the forward convolution).

\subsubsection{Gradient Propagation}
To compute the gradient propagation, the input gradient is convoluted with the 4D kernel. 
The data flow is the same as for the forward propagation since the input gradient is computed as the input feature in the forward computations.

\subsubsection{Dense Layer Computation}
This section illustrates how the dense layer operations are conducted using the same $9$ MACs of the PU.

\begin{itemize}[leftmargin=*]
    \item \textbf{Forward propagation}: It is implemented as a matrix multiplication between the flatted features of the previous layer with a 2D matrix. The weight matrix has a size ($m$,$n$) where $m$ is the number of input features and $n$ is the number of output features. Note that in the last dense layer, the output features' value is equal to the number of classes. This number, due to the CL setup, is not static and changes during the operation.
    The output of the forward algorithm is implemented according to \Cref{eq:y_n_dense}, which is described considering to have a 3D matrix with sizes ($I$, $J$, $K$) instead of the conventional flattened values.
    
    \begin{equation}
    \label{eq:y_n_dense}
    \begin{gathered}
    y_n = \sum_{i = 0}^{m} I_i \cdot W_{i,n} = \sum_{i,j,k=0}^{I,J,K} I_{i,j,k} \cdot W_{i,j,k,n}
    \end{gathered}
    \end{equation}
    
    In this way, we can reuse the same logic of the forward computation for convolutional layers. In this case, for each clock cycle, $8$ pixels of $8$ channels of the input feature and $8$ weights of $8$ channels are read and sent to $8$ of the $9$ MACs. All $64$ results are then summed together and saved into the partial sum register. This is done for the whole set of input features and it is repeated $n$ times, where $n$ is the number of output features. In this way, we can dynamically set the number of iterations required.
    
    \item \textbf{Gradient propagation}: The gradient propagation $dX$ is equal to the matrix multiplication between input gradient propagation and the transpose of the weights, as described in \Cref{eq:dX}. In this case, the optimization is more complicated. Due to the fact the size of $dY$ is dynamic due to the CL settings and is not necessarily a power of $2$, we cannot reach a $100\%$ utilization of the hardware resources. We propose to execute inside a single MAC the computation for each feature of $dX$, iteratively using the register and the partial sum logic. Using $9$ MACs, we can compute $9$ pixel in $n/8$ clock cycles, where $n$ is the number of output features, and the whole computation in $(I/9) \cdot (n/8)$ cycles. An example of how MACs are indexed is shown in \Cref{eq:MAC_0}.
    
    \begin{equation}
    \label{eq:MAC_0}
    MAC_{0,0} <= dX_{0} = \sum_{n=0}^{7}dY_n \cdot W^{T}_{n,0}
    \end{equation}
    
    The values will be saved in the partial sum register and summed in the next addition, until the end of the vector. 
    \item \textbf{Weight derivative}: The derivative of the weight ($dW$) is the matrix multiplication between the input feature $I$ used during the forward computation and the gradient propagation coming from the loss computation $dY$, as described in \Cref{eq:dW}.
    
    Using the matrix notation, since both operands are row matrix, their product produces a matrix. This means that the whole matrix $I$ is multiplied with the same value $dY_n$. This calculus can be executed in our MAC. $64$ parallel input features are read ($8$ pixels for $8$ channels, as in forward) and multiplied with one feature of $dY$. The 64 multiplication results are then added together and accumulated to previous multiplication and addition. This is repeated for each feature of $dY$.
\end{itemize}

\section{Experimental Results}
\label{sec:experiments}

\subsection{Experimental Setup}
\label{subsec:Exp_setup}

\begin{figure}[!ht]
\centering
\includegraphics[width=.99\linewidth]{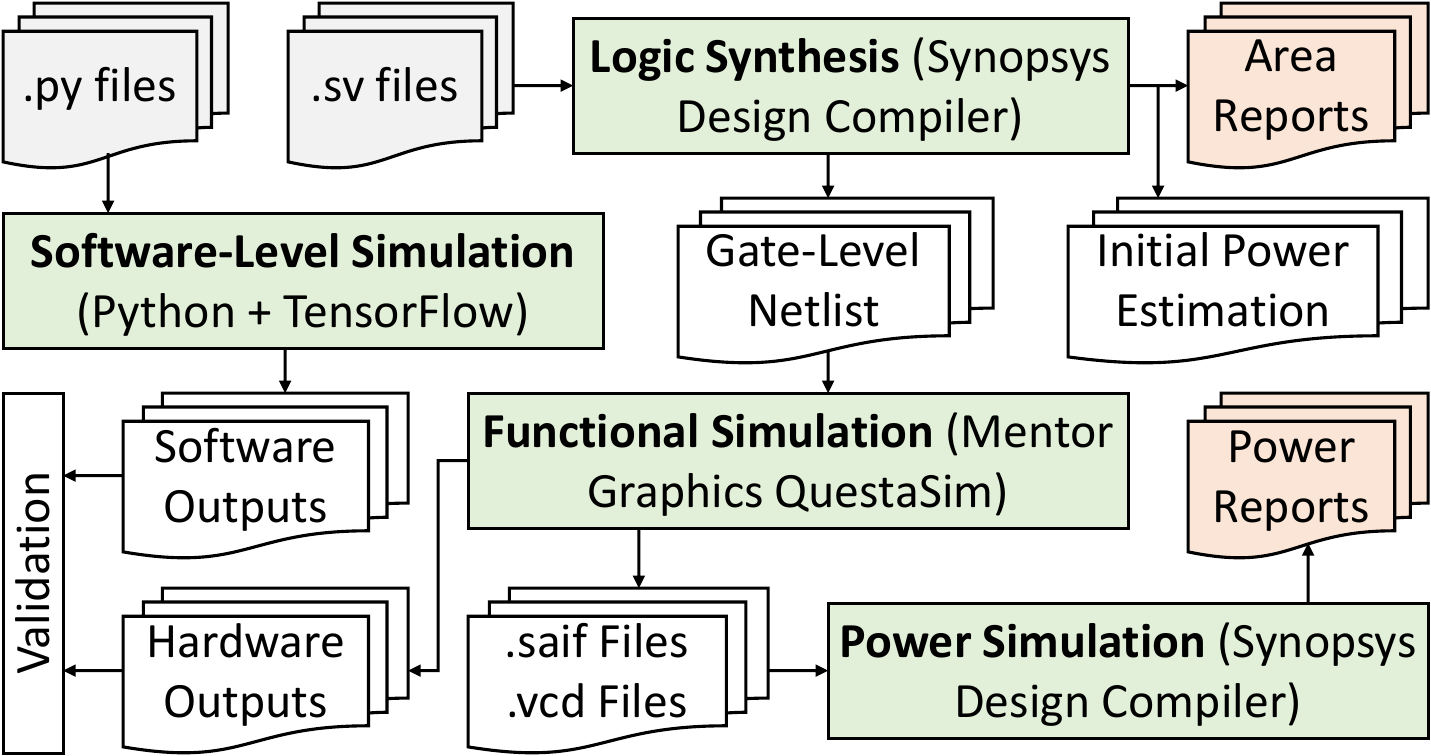}
\caption{Experimental setup and tool flow for conducting the experiments.}
\label{fig:Exp_setup}
\end{figure}

The complete TinyCL architecture is described at the RTL level using the SystemVerilog language. We synthesized the architecture in a 65 nm CMOS technology node with the ASIC design flow using the Synopsys Design Compiler tool. The synthesized netlist has been functionally tested through gate-level simulations, compared to the equivalent software-level implementation on TensorFlow~\cite{tensorflow} running on an NVIDIA Tesla P100 GPU. The complete flow of our experiments is shown in \Cref{fig:Exp_setup}. For comparison purposes, we test our system executing a simple DNN model, composed of 2 convolutional layers with ReLU activation, followed by a Dense layer. This model has been trained with the CIFAR10 dataset~\cite{cifar10} for 10 epochs, a learning rate of 1, and a batch size of 1, while the training is composed of 5 tasks of 2 classes per task, following the GDumb approach~\cite{Prabhu_2020ECCV_GDumb}. The model is used side by side with a memory of 6.144 MB, which can contain 1000 $32 \times 32$ RGB training samples. 

\subsection{Sysnthesis Evaluation of Area, Power, and Latency}
\label{subsec:evaluation}

\begin{figure}[!ht]
\centering
\includegraphics[width=.99\linewidth]{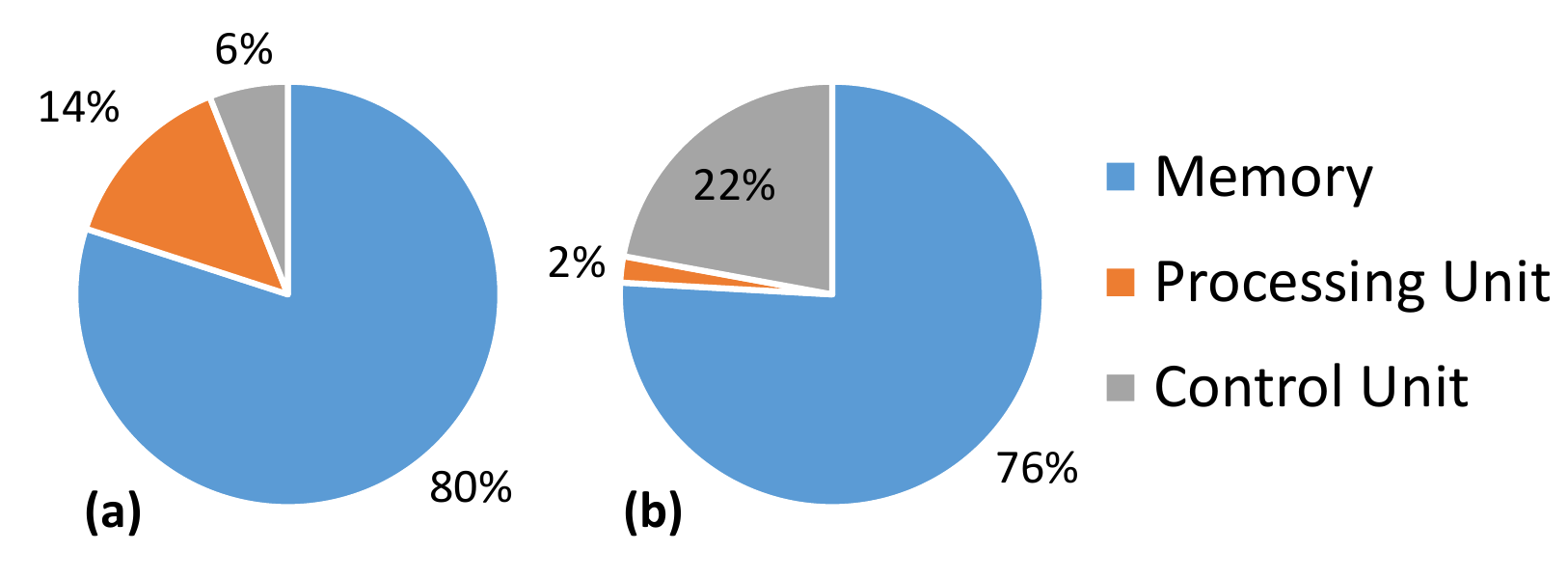}
\vspace*{-20pt}
\caption{(a) Area and (b) power breakdown of our TinyCL architecture.}
\label{fig:power_area_break}
\end{figure}

The synthesized TinyCL architecture has a clock period of 3.87 ns, consumes 86 mW, and occupies an area of 4.74 mm\textsuperscript{2}. \Cref{fig:power_area_break} analyzes in detail the power and area breakdown for each component. It is evident that the Memory block is responsible for the majority of area (80\%) and power (76\%) of the complete architecture.

Our architecture takes 8,192 clock cycles to compute either the forward convolution, the gradient propagation, or the gradient of the weight when we use 8 filters and the input feature has a shape of $32\times32\times8$. Instead, the dense layer for a feature of shape $32\times32\times8$ which outputs an array of size 10 takes 1,280 clock cycles for forward computation, 1,821 clock cycles for the computation of the gradients of the weights, and 1,280 clock cycles for the gradient propagation.

\subsection{Comparison with Related Works}
\label{subsec:comparison}

As a first analysis, we compare our TinyCL architecture with its software-level implementation on a P100 GPU. Following the setup described in \Cref{subsec:Exp_setup}, the GPU implementation lasts 103 s, while the TinyCL architecture executes the same workload in 1.76 s, thus reaching a speedup of $58 \times$.

\Cref{tab:results} compares latency, power, area, and performance of our proposed TinyCL architecture with related works. Note that since there are no available designs specialized for CL workloads, we compare it with existing designs that execute generic DNN training operations. The results show that the proposed TinyCL architecture is suitable for autonomous systems with limited resources since it achieves lower area and power consumption than prior works.

\begin{table}[!ht]
\centering
\vspace{10pt}
\caption{Comparison between TinyCL and related DNN Training architectures.}
\label{tab:results}
\begin{adjustbox}{max width=.99\linewidth}
\begin{tabular}{ |c|c|c|c|c|  }
 \hline
 Architecture & \makecell{Latency\\ (ns)} & \makecell{Power\\ (mW)} & \makecell{Area\\ (mm$^2$)} & \makecell{Performance\\ (TOPS)}\\
 \hline
 HNPU \cite{Han_2021JSSC_HNPU} & 4 & 1162& 12.96& 3.07\\
 LNPU \cite{Lee_2019ISSCC_LNPU}& 5 & 367&16 & 0.6\\
ISSCC19 \cite{Kim_2019ISSCC_MobileDepRLAccelerator} & 5 & 196 &16 &0.204\\
 \textbf{TinyCL (our)}   & \textbf{3.87}    &\textbf{86}&   \textbf{4.74}& \textbf{0.037}\\
 \hline
\end{tabular}
\end{adjustbox}
\end{table}

\section{Conclusion}

We have presented TinyCL, an efficient hardware architecture for CL algorithms, which executes all the involved operations such as forward propagation, backward propagation, and parameter update. A specialized control unit manages the memory and data flow to support memory-based CL policies and different operations on the same Processing Unit. The TinyCL architecture, synthesized in a 65 nm CMOS technology node, achieves $58 \times$ speedup compared to an equivalent software-level implementation executing on an Nvidia Tesla P100 GPU, while achieving lower area and power consumption than related works. Our design and analyses open new avenues for developing efficient CL algorithms on resource-constrained autonomous systems.

\section*{Acknowledgements}
This work was supported in part by the NYUAD Center for Cyber Security (CCS), funded by Tamkeen under the NYUAD Research Institute Award G1104.



% ...



\bibliographystyle{IEEEtran}
\bibliography{main.bib}

\begin{thebibliography}{10}
\providecommand{\url}[1]{#1}
\csname url@samestyle\endcsname
\providecommand{\newblock}{\relax}
\providecommand{\bibinfo}[2]{#2}
\providecommand{\BIBentrySTDinterwordspacing}{\spaceskip=0pt\relax}
\providecommand{\BIBentryALTinterwordstretchfactor}{4}
\providecommand{\BIBentryALTinterwordspacing}{\spaceskip=\fontdimen2\font plus
\BIBentryALTinterwordstretchfactor\fontdimen3\font minus \fontdimen4\font\relax}
\providecommand{\BIBforeignlanguage}[2]{{%
\expandafter\ifx\csname l@#1\endcsname\relax
\typeout{** WARNING: IEEEtran.bst: No hyphenation pattern has been}%
\typeout{** loaded for the language `#1'. Using the pattern for}%
\typeout{** the default language instead.}%
\else
\language=\csname l@#1\endcsname
\fi
#2}}
\providecommand{\BIBdecl}{\relax}
\BIBdecl

\bibitem{Chai_2021MLWA_DLComputerVision}
\BIBentryALTinterwordspacing
J.~Chai, H.~Zeng, A.~Li, and E.~W. Ngai, ``Deep learning in computer vision: A critical review of emerging techniques and application scenarios,'' \emph{Machine Learning with Applications}, 2021. [Online]. Available: \url{https://www.sciencedirect.com/science/article/pii/S2666827021000670}
\BIBentrySTDinterwordspacing

\bibitem{Chandani_2022IJDATS_MLFinance}
\BIBentryALTinterwordspacing
A.~Chandani and R.~Dubey, ``Application of machine learning in banking and finance: {A} bibliometric analysis,'' \emph{Int. J. Data Anal. Tech. Strateg.}, 2022. [Online]. Available: \url{https://doi.org/10.1504/ijdats.2022.10050641}
\BIBentrySTDinterwordspacing

\bibitem{An_2023Sensors_MLHelathcare}
\BIBentryALTinterwordspacing
Q.~An, S.~Rahman, J.~Zhou, and J.~J. Kang, ``A comprehensive review on machine learning in healthcare industry: Classification, restrictions, opportunities and challenges,'' \emph{Sensors}, 2023. [Online]. Available: \url{https://doi.org/10.3390/s23094178}
\BIBentrySTDinterwordspacing

\bibitem{Soori_2023COGR_DLRobotics}
\BIBentryALTinterwordspacing
M.~Soori, B.~Arezoo, and R.~Dastres, ``Artificial intelligence, machine learning and deep learning in advanced robotics, a review,'' \emph{Cognitive Robotics}, 2023. [Online]. Available: \url{https://www.sciencedirect.com/science/article/pii/S2667241323000113}
\BIBentrySTDinterwordspacing

\bibitem{Capra_2020Access_SurveyDNNs}
\BIBentryALTinterwordspacing
M.~Capra, B.~Bussolino, A.~Marchisio, G.~Masera, M.~Martina, and M.~Shafique, ``Hardware and software optimizations for accelerating deep neural networks: Survey of current trends, challenges, and the road ahead,'' \emph{{IEEE} Access}, 2020. [Online]. Available: \url{https://doi.org/10.1109/ACCESS.2020.3039858}
\BIBentrySTDinterwordspacing

\bibitem{Dhilleswararao_2022Access_EfficientDL}
\BIBentryALTinterwordspacing
P.~Dhilleswararao, S.~Boppu, M.~S. Manikandan, and L.~R. Cenkeramaddi, ``Efficient hardware architectures for accelerating deep neural networks: Survey,'' \emph{{IEEE} Access}, 2022. [Online]. Available: \url{https://doi.org/10.1109/ACCESS.2022.3229767}
\BIBentrySTDinterwordspacing

\bibitem{marchisio2021feeca}
\BIBentryALTinterwordspacing
A.~Marchisio, V.~Mrazek, M.~A. Hanif, and M.~Shafique, ``{FEECA:} design space exploration for low-latency and energy-efficient capsule network accelerators,'' \emph{IEEE Transactions on Very Large Scale Integration (VLSI) Systems}, vol.~29, no.~4, pp. 716--729, 2021. [Online]. Available: \url{https://doi.org/10.1109/TVLSI.2021.3059518}
\BIBentrySTDinterwordspacing

\bibitem{marchisio2023swifttron}
\BIBentryALTinterwordspacing
A.~Marchisio, D.~Dura, M.~Capra, M.~Martina, G.~Masera, and M.~Shafique, ``Swifttron: An efficient hardware accelerator for quantized transformers,'' in \emph{2023 International Joint Conference on Neural Networks (IJCNN)}.\hskip 1em plus 0.5em minus 0.4em\relax IEEE, 2023, pp. 1--9. [Online]. Available: \url{https://doi.org/10.1109/IJCNN54540.2023.10191521}
\BIBentrySTDinterwordspacing

\bibitem{putra2024snn4agents}
\BIBentryALTinterwordspacing
R.~V.~W. Putra, A.~Marchisio, and M.~Shafique, ``Snn4agents: a framework for developing energy-efficient embodied spiking neural networks for autonomous agents,'' \emph{Frontiers in Robotics and AI}, vol.~11, p. 1401677, 2024. [Online]. Available: \url{https://www.frontiersin.org/journals/robotics-and-ai/articles/10.3389/frobt.2024.1401677}
\BIBentrySTDinterwordspacing

\bibitem{Wang_2023arxiv_CLSurvey}
\BIBentryALTinterwordspacing
L.~Wang, X.~Zhang, H.~Su, and J.~Zhu, ``A comprehensive survey of continual learning: Theory, method and application,'' \emph{CoRR}, vol. abs/2302.00487, 2023. [Online]. Available: \url{https://doi.org/10.48550/arXiv.2302.00487}
\BIBentrySTDinterwordspacing

\bibitem{aswani2024examining}
\BIBentryALTinterwordspacing
N.~S. Aswani, A.~Guesmi, M.~A. Hanif, and M.~Shafique, ``Examining changes in internal representations of continual learning models through tensor decomposition,'' \emph{arXiv preprint arXiv:2405.03244}, 2024. [Online]. Available: \url{https://arxiv.org/abs/2405.03244}
\BIBentrySTDinterwordspacing

\bibitem{minhas2024continual}
\BIBentryALTinterwordspacing
M.~F. Minhas, R.~V.~W. Putra, F.~Awwad, O.~Hasan, and M.~Shafique, ``Continual learning with neuromorphic computing: Theories, methods, and applications,'' \emph{arXiv preprint arXiv:2410.09218}, 2024. [Online]. Available: \url{https://arxiv.org/abs/2410.09218}
\BIBentrySTDinterwordspacing

\bibitem{Mccloskey_1989Elsevier_CatastrophicForgetting}
\BIBentryALTinterwordspacing
M.~McCloskey and N.~J. Cohen, ``Catastrophic interference in connectionist networks: The sequential learning problem,'' in \emph{Psychology of Learning and Motivation}, G.~H. Bower, Ed., 1989. [Online]. Available: \url{https://www.sciencedirect.com/science/article/pii/S0079742108605368}
\BIBentrySTDinterwordspacing

\bibitem{Chen_2020Engineering_SurveyDNNAccelerators}
\BIBentryALTinterwordspacing
Y.~Chen, Y.~Xie, L.~Song, F.~Chen, and T.~Tang, ``A survey of accelerator architectures for deep neural networks,'' \emph{Engineering}, 2020. [Online]. Available: \url{https://www.sciencedirect.com/science/article/pii/S2095809919306356}
\BIBentrySTDinterwordspacing

\bibitem{Capra_2020MDPI_UpdatedSurveyDNN}
\BIBentryALTinterwordspacing
M.~Capra, B.~Bussolino, A.~Marchisio, M.~Shafique, G.~Masera, and M.~Martina, ``An updated survey of efficient hardware architectures for accelerating deep convolutional neural networks,'' \emph{Future Internet}, 2020. [Online]. Available: \url{https://doi.org/10.3390/fi12070113}
\BIBentrySTDinterwordspacing

\bibitem{Lee_2021SSCS_HWAcceleratorTraining}
\BIBentryALTinterwordspacing
J.~Lee and H.-J. Yoo, ``An overview of energy-efficient hardware accelerators for on-device deep-neural-network training,'' \emph{IEEE Open Journal of the Solid-State Circuits Society}, 2021. [Online]. Available: \url{https://doi.org/10.1109/OJSSCS.2021.3119554}
\BIBentrySTDinterwordspacing

\bibitem{Thrun_1995RAS_LifelongRobotLearning}
\BIBentryALTinterwordspacing
S.~Thrun and T.~M. Mitchell, ``Lifelong robot learning,'' \emph{Robotics Auton. Syst.}, 1995. [Online]. Available: \url{https://doi.org/10.1016/0921-8890(95)00004-Y}
\BIBentrySTDinterwordspacing

\bibitem{Knoblauch_2020ICML_ContinualLearningNPhard}
\BIBentryALTinterwordspacing
J.~Knoblauch, H.~Husain, and T.~Diethe, ``Optimal continual learning has perfect memory and is np-hard,'' in \emph{37th International Conference on Machine Learning (ICML)}, 2020. [Online]. Available: \url{http://proceedings.mlr.press/v119/knoblauch20a.html}
\BIBentrySTDinterwordspacing

\bibitem{Kirkpatrick_2016arxiv_CatastrophicForgettingEWC}
\BIBentryALTinterwordspacing
J.~Kirkpatrick \emph{et~al.}, ``Overcoming catastrophic forgetting in neural networks,'' \emph{CoRR}, vol. abs/1612.00796, 2016. [Online]. Available: \url{http://arxiv.org/abs/1612.00796}
\BIBentrySTDinterwordspacing

\bibitem{Li_2018TPAMI_LwF}
\BIBentryALTinterwordspacing
Z.~Li and D.~Hoiem, ``Learning without forgetting,'' \emph{{IEEE} Trans. Pattern Anal. Mach. Intell.}, 2018. [Online]. Available: \url{https://doi.org/10.1109/TPAMI.2017.2773081}
\BIBentrySTDinterwordspacing

\bibitem{LopezPaz_2017NeurIPS_GEM}
\BIBentryALTinterwordspacing
D.~Lopez{-}Paz and M.~Ranzato, ``Gradient episodic memory for continual learning,'' in \emph{Annual Conference on Neural Information Processing Systems (NeurIPS)}, 2017. [Online]. Available: \url{https://proceedings.neurips.cc/paper/2017/hash/f87522788a2be2d171666752f97ddebb-Abstract.html}
\BIBentrySTDinterwordspacing

\bibitem{Chaudhry_2019ICLR_AGEM}
\BIBentryALTinterwordspacing
A.~Chaudhry, M.~Ranzato, M.~Rohrbach, and M.~Elhoseiny, ``Efficient lifelong learning with {A-GEM},'' in \emph{7th International Conference on Learning Representations (ICLR)}, 2019. [Online]. Available: \url{https://openreview.net/forum?id=Hkf2\_sC5FX}
\BIBentrySTDinterwordspacing

\bibitem{Rebuffi_2017CVPR_iCaRL}
\BIBentryALTinterwordspacing
S.~Rebuffi, A.~Kolesnikov, G.~Sperl, and C.~H. Lampert, ``icarl: Incremental classifier and representation learning,'' in \emph{{IEEE} Conference on Computer Vision and Pattern Recognition (CVPR)}, 2017. [Online]. Available: \url{https://doi.org/10.1109/CVPR.2017.587}
\BIBentrySTDinterwordspacing

\bibitem{Hayes_2019ICRA_ExperienceReplay}
\BIBentryALTinterwordspacing
T.~L. Hayes, N.~D. Cahill, and C.~Kanan, ``Memory efficient experience replay for streaming learning,'' in \emph{International Conference on Robotics and Automation (ICRA)}, 2019. [Online]. Available: \url{https://doi.org/10.1109/ICRA.2019.8793982}
\BIBentrySTDinterwordspacing

\bibitem{Aljundi_2019NeurIPS_MIR}
\BIBentryALTinterwordspacing
R.~Aljundi \emph{et~al.}, ``Online continual learning with maximal interfered retrieval,'' in \emph{Annual Conference on Neural Information Processing Systems (NeurIPS)}, 2019. [Online]. Available: \url{https://proceedings.neurips.cc/paper/2019/hash/15825aee15eb335cc13f9b559f166ee8-Abstract.html}
\BIBentrySTDinterwordspacing

\bibitem{Aljundi_2019NeurIPS_GSS}
\BIBentryALTinterwordspacing
R.~Aljundi, M.~Lin, B.~Goujaud, and Y.~Bengio, ``Gradient based sample selection for online continual learning,'' in \emph{Annual Conference on Neural Information Processing Systems (NeurIPS)}, 2019. [Online]. Available: \url{https://proceedings.neurips.cc/paper/2019/hash/e562cd9c0768d5464b64cf61da7fc6bb-Abstract.html}
\BIBentrySTDinterwordspacing

\bibitem{Prabhu_2020ECCV_GDumb}
\BIBentryALTinterwordspacing
A.~Prabhu, P.~H.~S. Torr, and P.~K. Dokania, ``Gdumb: {A} simple approach that questions our progress in continual learning,'' in \emph{16th European Conference on Computer Vision (ECCV)}, 2020. [Online]. Available: \url{https://doi.org/10.1007/978-3-030-58536-5\_31}
\BIBentrySTDinterwordspacing

\bibitem{Shin_2017NeurIPS_DGR}
\BIBentryALTinterwordspacing
H.~Shin, J.~K. Lee, J.~Kim, and J.~Kim, ``Continual learning with deep generative replay,'' in \emph{Annual Conference on Neural Information Processing Systems (NeurIPS)}, 2017. [Online]. Available: \url{https://proceedings.neurips.cc/paper/2017/hash/0efbe98067c6c73dba1250d2beaa81f9-Abstract.html}
\BIBentrySTDinterwordspacing

\bibitem{Lee_2020ICLR_CNDPM}
\BIBentryALTinterwordspacing
S.~Lee, J.~Ha, D.~Zhang, and G.~Kim, ``A neural dirichlet process mixture model for task-free continual learning,'' in \emph{8th International Conference on Learning Representations (ICLR)}, 2020. [Online]. Available: \url{https://openreview.net/forum?id=SJxSOJStPr}
\BIBentrySTDinterwordspacing

\bibitem{Du_2019ICMLA_PST}
\BIBentryALTinterwordspacing
X.~Du, G.~Charan, F.~Liu, and Y.~Cao, ``Single-net continual learning with progressive segmented training,'' in \emph{18th {IEEE} International Conference On Machine Learning And Applications (ICMLA)}, 2019. [Online]. Available: \url{https://doi.org/10.1109/ICMLA.2019.00267}
\BIBentrySTDinterwordspacing

\bibitem{Venkataramani_2017ISCA_ScaleDeep}
\BIBentryALTinterwordspacing
S.~Venkataramani \emph{et~al.}, ``Scaledeep: {A} scalable compute architecture for learning and evaluating deep networks,'' in \emph{44th Annual International Symposium on Computer Architecture (ISCA)}, 2017. [Online]. Available: \url{https://doi.org/10.1145/3079856.3080244}
\BIBentrySTDinterwordspacing

\bibitem{Kim_2018TCAD_DeepTrain}
\BIBentryALTinterwordspacing
D.~Kim, T.~Na, S.~Yalamanchili, and S.~Mukhopadhyay, ``Deeptrain: {A} programmable embedded platform for training deep neural networks,'' \emph{{IEEE} Trans. Comput. Aided Des. Integr. Circuits Syst.}, 2018. [Online]. Available: \url{https://doi.org/10.1109/TCAD.2018.2858358}
\BIBentrySTDinterwordspacing

\bibitem{Jain_2018ISCA_Gist}
\BIBentryALTinterwordspacing
A.~Jain, A.~Phanishayee, J.~Mars, L.~Tang, and G.~Pekhimenko, ``Gist: Efficient data encoding for deep neural network training,'' in \emph{45th {ACM/IEEE} Annual International Symposium on Computer Architecture (ISCA)}, 2018. [Online]. Available: \url{https://doi.org/10.1109/ISCA.2018.00070}
\BIBentrySTDinterwordspacing

\bibitem{Qin_2020HPCA_SIGMA}
\BIBentryALTinterwordspacing
E.~Qin \emph{et~al.}, ``{SIGMA:} {A} sparse and irregular {GEMM} accelerator with flexible interconnects for {DNN} training,'' in \emph{{IEEE} International Symposium on High Performance Computer Architecture (HPCA)}, 2020. [Online]. Available: \url{https://doi.org/10.1109/HPCA47549.2020.00015}
\BIBentrySTDinterwordspacing

\bibitem{Yang_2020MICRO_Procrustes}
\BIBentryALTinterwordspacing
D.~Yang \emph{et~al.}, ``Procrustes: a dataflow and accelerator for sparse deep neural network training,'' in \emph{53rd Annual {IEEE/ACM} International Symposium on Microarchitecture (MICRO)}, 2020. [Online]. Available: \url{https://doi.org/10.1109/MICRO50266.2020.00064}
\BIBentrySTDinterwordspacing

\bibitem{Lee_2019ISSCC_LNPU}
\BIBentryALTinterwordspacing
J.~Lee, J.~Lee, D.~Han, J.~Lee, G.~Park, and H.-J. Yoo, ``7.7 lnpu: A 25.3 tflops/w sparse deep-neural-network learning processor with fine-grained mixed precision of fp8-fp16,'' in \emph{2019 IEEE International Solid-State Circuits Conference-(ISSCC)}, 2019. [Online]. Available: \url{https://doi.org/10.1109/ISSCC.2019.8662302}
\BIBentrySTDinterwordspacing

\bibitem{Han_2021JSSC_HNPU}
\BIBentryALTinterwordspacing
D.~Han, D.~Im, G.~Park, Y.~Kim, S.~Song, J.~Lee, and H.~Yoo, ``{HNPU:} an adaptive {DNN} training processor utilizing stochastic dynamic fixed-point and active bit-precision searching,'' \emph{{IEEE} J. Solid State Circuits}, 2021. [Online]. Available: \url{https://doi.org/10.1109/JSSC.2021.3066400}
\BIBentrySTDinterwordspacing

\bibitem{Noh_2023TC_FlexBlock}
\BIBentryALTinterwordspacing
S.~Noh, J.~Koo, S.~Lee, J.~Park, and J.~Kung, ``Flexblock: {A} flexible {DNN} training accelerator with multi-mode block floating point support,'' \emph{{IEEE} Trans. Computers}, 2023. [Online]. Available: \url{https://doi.org/10.1109/TC.2023.3253050}
\BIBentrySTDinterwordspacing

\bibitem{Lu_2023TNNLS_ETA}
\BIBentryALTinterwordspacing
J.~Lu, C.~Ni, and Z.~Wang, ``{ETA:} an efficient training accelerator for dnns based on hardware-algorithm co-optimization,'' \emph{{IEEE} Trans. Neural Networks Learn. Syst.}, 2023. [Online]. Available: \url{https://doi.org/10.1109/TNNLS.2022.3145850}
\BIBentrySTDinterwordspacing

\bibitem{Kim_2019ISSCC_MobileDepRLAccelerator}
\BIBentryALTinterwordspacing
C.~Kim, S.~Kang, D.~Shin, S.~Choi, Y.~Kim, and H.~Yoo, ``A 2.1tflops/w mobile deep {RL} accelerator with transposable {PE} array and experience compression,'' in \emph{{IEEE} International Solid- State Circuits Conference (ISSCC)}, 2019. [Online]. Available: \url{https://doi.org/10.1109/ISSCC.2019.8662447}
\BIBentrySTDinterwordspacing

\bibitem{Huai_2023FGCS_ZerorizedBatchNormalization}
\BIBentryALTinterwordspacing
S.~Huai, D.~Liu, H.~Kong, W.~Liu, R.~Subramaniam, C.~Makaya, and Q.~Lin, ``Latency-constrained {DNN} architecture learning for edge systems using zerorized batch normalization,'' \emph{Future Gener. Comput. Syst.}, 2023. [Online]. Available: \url{https://doi.org/10.1016/j.future.2022.12.021}
\BIBentrySTDinterwordspacing

\bibitem{Wang_2022NeurIPS_SparCL}
\BIBentryALTinterwordspacing
Z.~Wang \emph{et~al.}, ``Sparcl: Sparse continual learning on the edge,'' in \emph{Annual Conference on Neural Information Processing Systems (NeurIPS)}, 2022. [Online]. Available: \url{http://papers.nips.cc/paper\_files/paper/2022/hash/80133d0f6eccaace15508f91e3c5a93c-Abstract-Conference.html}
\BIBentrySTDinterwordspacing

\bibitem{Harun_2023arxiv_SIESTA}
\BIBentryALTinterwordspacing
M.~Y. Harun, J.~Gallardo, T.~L. Hayes, R.~Kemker, and C.~Kanan, ``{SIESTA:} efficient online continual learning with sleep,'' \emph{CoRR}, vol. abs/2303.10725, 2023. [Online]. Available: \url{https://doi.org/10.48550/arXiv.2303.10725}
\BIBentrySTDinterwordspacing

\bibitem{Ma_2023MobiCom_Miro}
\BIBentryALTinterwordspacing
X.~Ma, S.~Jeong, M.~Zhang, D.~Wang, J.~Choi, and M.~Jeon, ``Cost-effective on-device continual learning over memory hierarchy with miro,'' in \emph{29th Annual International Conference on Mobile Computing and Networking (MobiCom)}, 2023. [Online]. Available: \url{https://doi.org/10.1145/3570361.3613297}
\BIBentrySTDinterwordspacing

\bibitem{gaur2020training}
\BIBentryALTinterwordspacing
D.~Gaur, J.~Folz, and A.~Dengel, ``Training deep neural networks without batch normalization,'' \emph{CoRR}, vol. abs/2008.07970, 2020. [Online]. Available: \url{https://arxiv.org/abs/2008.07970}
\BIBentrySTDinterwordspacing

\bibitem{tensorflow}
\BIBentryALTinterwordspacing
M.~Abadi \emph{et~al.}, ``Tensorflow: {A} system for large-scale machine learning,'' in \emph{12th {USENIX} Symposium on Operating Systems Design and Implementation (OSDI)}, 2016. [Online]. Available: \url{https://www.usenix.org/conference/osdi16/technical-sessions/presentation/abadi}
\BIBentrySTDinterwordspacing

\bibitem{cifar10}
\BIBentryALTinterwordspacing
A.~Krizhevsky, V.~Nair, and G.~Hinton, ``Cifar-10 (canadian institute for advanced research),'' 2015. [Online]. Available: \url{http://www.cs.toronto.edu/~kriz/cifar.html}
\BIBentrySTDinterwordspacing

\end{thebibliography}

\end{document}